\documentclass[a4paper,conference]{IEEEtran}

\ifCLASSINFOpdf
\else
\fi
\usepackage{times}
\usepackage{epsfig}
\usepackage{graphicx}
\usepackage{amsmath}
\usepackage{amssymb}
\usepackage{multirow}
\usepackage{subfigure}
\usepackage{xfrac}
\usepackage{float}
\usepackage{booktabs}

\def\etal{\emph{et al. }}

\begin{document}

\title{Learning Stereo Matchability in\\Disparity Regression Networks}

\author{
\IEEEauthorblockN{Jingyang Zhang$^1$, Yao Yao$^1$, Zixin Luo$^1$, Shiwei Li$^2$, Tianwei Shen$^1$, Tian Fang$^2$, Long Quan$^1$}
\IEEEauthorblockA{
\begin{tabular}{cc}
	\begin{tabular}[c]{@{}c@{}}$^1$The Hong Kong University of Science and Technology\\ Hong Kong SAR, China\\ Email: \{jzhangbs,yyaoag,zluoag,tshenaa,quan\}@cse.ust.hk\end{tabular} & \begin{tabular}[c]{@{}c@{}}$^2$Everest Innovation Technology Ltd.\\ Hong Kong SAR, China\\ Email: \{sli,fangtian\}@altizure.com\end{tabular}
\end{tabular}}
}


%


\maketitle

\begin{abstract}
Learning-based stereo matching has recently achieved promising results, yet still suffers difficulties in establishing reliable matches in weakly matchable regions that are textureless, non-Lambertian, or occluded. In this paper, we address this challenge by proposing a stereo matching network that considers pixel-wise matchability. Specifically, the network jointly regresses disparity and matchability maps from 3D probability volume through expectation and entropy operations. Next, a learned attenuation is applied as the robust loss function to alleviate the influence of weakly matchable pixels in the training. Finally, a matchability-aware disparity refinement is introduced to improve the depth inference in weakly matchable regions. The proposed deep stereo matchability (DSM) framework can improve the matching result or accelerate the computation while still guaranteeing the quality. Moreover, the DSM framework is portable to many recent stereo networks. Extensive experiments are conducted on Scene Flow and KITTI stereo datasets to demonstrate the effectiveness of the proposed framework over the state-of-the-art learning-based stereo methods. 
\end{abstract}


%
\IEEEpeerreviewmaketitle

\section{Introduction}
Stereo matching recovers dense correspondences between an image pair, which is widely applied to 3D mapping, scene understanding and autonomous driving. While traditional methods apply hand-crafted matching costs and engineered regularization for stereo matching, recent learning-based methods formulate the problem into an end-to-end trainable task. Specifically, a typical deep stereo architecture \cite{kendall2017end,chang2018pyramid,cheng2019learning,zhang2019ga} includes four parts: a) image feature maps are extracted from original images via 2D Convolutional Neural Networks (CNNs); b) a 3D cost volume is built from two feature maps; c) a 3D probability volume is regularized from the 3D cost volume through 3D CNNs; d) the final disparity map is regressed from the 3D probability volume via dimension reduction such as \textit{soft-argmin} operation \cite{kendall2017end}. Compared with traditional methods, deep stereo matching networks benefit from larger receptive fields and exploit global context information from images, which significantly boosts the matching accuracy over traditional methods on recent stereo benchmarks \cite{geiger2012we,menze2015object,mayer2016large}.

Despite the recent achievements,
it is still challenging to recover dense correspondences in textureless, non-Lambertian or occluded regions, where it is often the case that only few or even no visual correspondence can be reliably established across different views. To some extent, such regions are unmatchable from visual contents. However, most existing stereo matching networks do not make an explicit identification and evenly matched them with other pixels in the image, which leads to wrong disparity values in unmatchable regions and deteriorate the training process. 

Ideally, the matching algorithm is supposed to be capable of specializing different strategies on pixels of different matchability. Matchability (also referred to as confidence) estimation has long been studied in stereo vision, whereas previous methods are mostly targeted at classical stereo matching \cite{hu2012quantitative} or learning-based patch-wise stereo matching \cite{poggi2017quantitative}. Recently, Kendall and Gal~\cite{kendall2017uncertainties} proposed an uncertainty estimation framework based on Bayesian deep learning, where the output and its uncertainty are jointly retrieved from the last layer feature maps. Their framework is suitable for single-image vision tasks such as 2D semantic segmentation and monocular depth map estimation. However, in the context of deep stereo matching, the matchability should be retrieved from 3D probability volume rather than 2D feature maps, and no such mapping rules has been proposed.

In this paper, we address the importance of pixel-wise matchability for learning-based stereo matching.
Specifically, we propose a method that estimates the matchability from the probability volume via the disparity-wise \textit{entropy}, which models the probability distribution to interpret the disparity quality. With the estimated matchability, we derive a robust loss function to reduce the negative influence of unmatchable pixels during the training process based on the Laplacian distribution assumption \cite{kendall2017uncertainties}. Moreover, the matchability provides valuable information regarding unreliable matching regions. We further exploit the matchability map and input image context to refine the disparity especially for weakly matchable regions. 

Intuitively, the matchability is also related to the uncertainty or confidence of the estimation. These concepts were extensively studied in previous works but lack unified definitions. Here we would like to clarify the definitions of three conceptually similar terms \textit{probability}, \textit{matchability} and \textit{uncertainty} used in this paper. a) The \textit{probability} is always associated with the 3D volume, which indicates the probability value of the disparity lying at each voxel, and is yielded from the initial cost volume via 3D CNNs; b) the \textit{matchability} is a 2D map computed from the probability volume via disparity-wise \textit{entropy} operation. This concept is also referred to \textit{confidence} in previous literatures; c) the \textit{uncertainty} is computed from the \textit{matchability} via a learnable mapping, which is used to weight the training loss for joint disparity and matchability training (see Sec.~\ref{sec:loss}). 

The proposed deep stereo matchability (DSM) framework can be applied to most of the current state-of-the-art stereo networks \cite{kendall2017end,chang2018pyramid,cheng2019learning,zhang2019ga} and consistently boosts their performance. Also, we find that our method can still  guarantee a good disparity estimation quality while reducing the expensive 3D convolution layers, which makes the method suitable for real-time applications or platforms with limited computation power. We test our methods on the Scene Flow \cite{mayer2016large}, KITTI 2012 \cite{geiger2012we} and KITTI 2015 \cite{menze2015object} stereo benchmarks. Our main contributions are summarized below.
\begin{itemize}
	\item We propose a unified stereo matchability framework, which is applicable to most of the state-of-the-art stereo matching networks.
	\item We introduce a disparity-wise \textit{entropy} operation to retrieve the matchability map from the 3D probability volume.
	\item We introduce a matchability-aware disparity refinement network to enhance unmatchable regions by exploiting contextual information from both the input image and the matchability map.
\end{itemize}

\begin{figure*}[t!]
	\centering
	\includegraphics[width=1\linewidth]{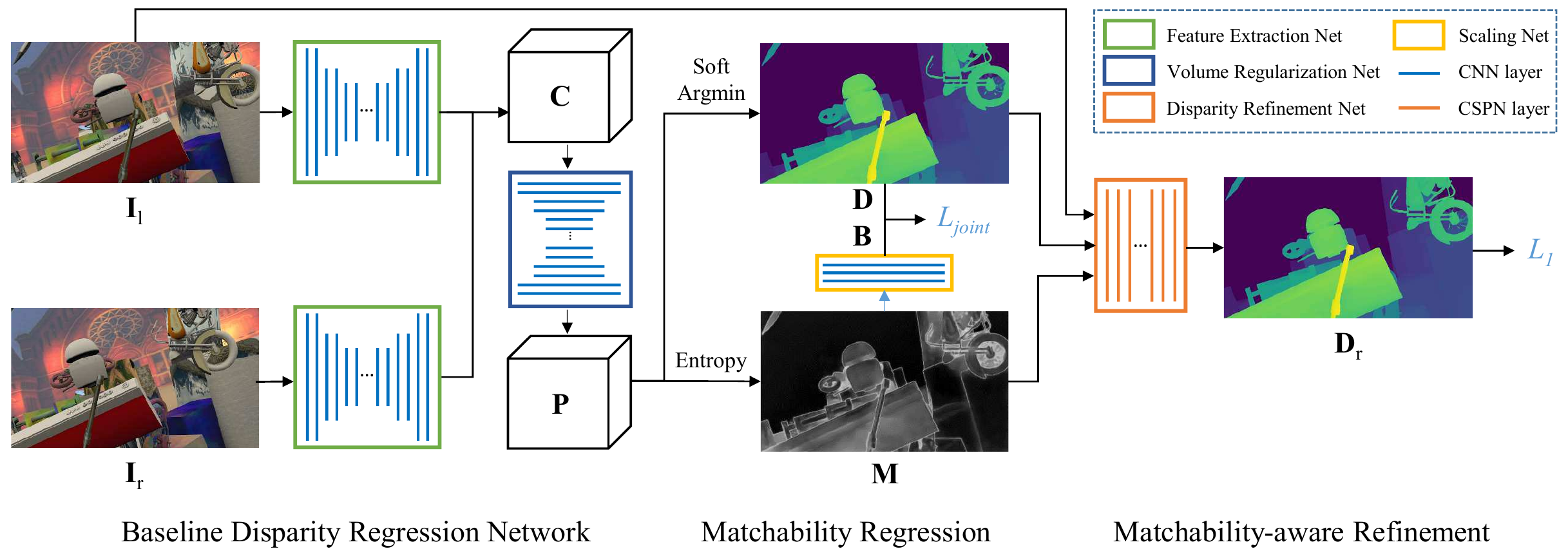}
	\vspace{-6mm}
	\caption{The proposed framework. Our network contains a baseline disparity regression network, where the image features are extracted though a 2D UNet and the cost volume are regularized via a 3D UNet. The matchability and the initial disparity maps are respectively regressed from the probability volume using the \textit{soft-argmin} and the \textit{entropy} operations. Finally we use the matchability information and input image semantics to refine the disparity output.}
	\label{fig:framework}
	\vspace{-3mm}
\end{figure*}


\section{Related Work}
\paragraph{Learning-based Stereo}
Deep learning has demonstrated its excellent performance for disparity estimation and has gradually been applied to each component of the stereo reconstruction pipeline. 
In contrast to hand-crafted image features and matching costs \cite{hirschmuller2007evaluation}, deep image features and learned cost metrics \cite{han2015matchnet,zbontar2016stereo,luo2016efficient} are proposed for pair-wise pixel matching. Later on, researchers also apply learning techniques to cost volume regularization. SGMNet \cite{seki2017sgm} proposes to learn parameters for the classical semi-global matching regularization \cite{hirschmuller2008stereo}, and CNN-CRF \cite{knobelreiter2017end} uses the network to model the Conditional Random Field optimization. Further, FlowNet \cite{dosovitskiy2015flownet} and DispNet \cite{mayer2016large} introduce fully learnable stereo architectures to regress the disparity map from the cost volume through 2D CNNs. 

Recently, Kendall \etal propose a popular disparity regression network called GCNet \cite{kendall2017end}. Their network constructs a cost volume from deep image features and applies 3D CNNs to regularize the volume. The soft argmin operation is introduced to regress the disparity map from the probability volume. This disparity regression framework is adopted by many recent stereo networks. For example, PSMNet \cite{chang2018pyramid} exploits the multi-scale context information for both image feature extraction and cost volume regularization; CSPN \cite{cheng2019learning} introduces the convolutional spatial propagation network (CSPN) to further refine the depth/disparity output. GANet \cite{zhang2019ga} models the semi-global matching process in their cost volume regularization network. GwcNet \cite{guo2019group} applies group-wise correlation instead of concatenation in the cost volume construction to reduce the memory consumption while still guarantee the performance. SegStereo \cite{yang2018SegStereo} and EdgeStereo \cite{song2020edgestereo} take edge and segmentation information into account. These methods have achieved state-of-the-art performances on Scene Flow \cite{mayer2016large} and KITTI stereo benchmarks \cite{geiger2012we,menze2015object}. In this paper, we target to learn the stereo matchability especially for such disparity regression networks. 

\paragraph{Matchability Estimation}
Previous works on classical stereo matchability have been extensively reviewed by Hu and Mordohai \cite{hu2012quantitative}, and we refer readers to this paper for more comprehensive review. More recently, CNN-based approaches are also proposed for matchability or confidence estimation \cite{poggi2017quantitative}. Researchers first applied learning-based method to local patch-wise matchability/confidence estimation \cite{poggi2017learning,seki2016patch,poggi2016learning}, and then used deep neural network for end-to-end confidence map inference from multi-modal input \cite{poggi2016learning,kim2018unified,tosi2018beyond,kim2019laf}. These methods usually generate the confidence map as network output, however, fail to utilize the confidence information to further improve the disparity estimation.

Recently, an uncertainty estimation framework \cite{kendall2017uncertainties} is proposed based on Bayesian deep learning to jointly retrieve the network output and its uncertainty. However, this framework is retricted to image level tasks (e.g., 2D semantic segmentation). For the aforementioned deep stereo networks \cite{kendall2017end,cheng2019learning}, we need to regress the matchability from the 3D probability volume, and it is still not clear what is the best practice to retrieve such information. To this end, we will introduce the disparity-wise $entropy$ operation to regress the matchability for each image pixel.

\paragraph{Disparity Refinement}

Due to the ill-posedness of the disparity estimation on unmatchable pixels, a post-processing module is often employed to improve the estimated details from an initial disparity map. Traditionally, this problem is approached by such as bilateral image filtering~\cite{barron2016fast} or data-driven methods through total variation (TV)~\cite{ferstl2013image}. More recently, statistical modeling methods are introduced such as the conditional random filed (CRF)~\cite{wang2015towards} or its learning-based variants~\cite{zheng2015conditional} to allow an effective joint training. Meanwhile, another line of works seek to eschew the complex heuristics through careful network designs~\cite{xu2017deep,yao2018mvsnet,cheng2019learning}. In particular, Convolutional Spatial Propagation Network (CSPN) \cite{cheng2019learning} has demonstrated appealing results by an an-isotropic diffusion simulation, so as to convolve each pixel with different kernels for disparity enhancement. In this paper, we collaboratively equip the matchability-awareness and CSPN in the refinement network, and target to improve the overall estimation accuracy in an end-to-end fashion.


\section{Method}
In this section, we present the proposed stereo matchability framework. We first describe our baseline model for disparity regression in (Sec.~\ref{sec:baseline}). Then, we introduce the disparity-wise entropy operation to regress the matchability map from the 3D probability volume (Sec.~\ref{sec:entropy}). The disparity and the mapping from matchability to uncertainty are jointly trained with a robust loss function by assuming the disparity output follows the Laplacian distribution (Sec.~\ref{sec:loss}). Lastly, the disparity refinement network is introduced to further refine the disparity map with the semantic information from the input image and the estimated matchability map (Sec.~\ref{sec:refinement}).

\subsection{Disparity Regression Network}\label{sec:baseline}

We follow the recent best stereo practices \cite{kendall2017end,chang2018pyramid,cheng2019learning} to construct our baseline network for disparity regression. Given a stereo image pair $\{\mathbf{I}_l, \mathbf{I}_r\}$ of size $H \times W$, we first apply a siamese 2D convolutional neural network to extract unary features $\{\mathbf{F}_l, \mathbf{F}_r\}$ of size $\frac{H}{4} \times \frac{W}{4} \times C$ for both images. Next, unary image features are used to construct a cost volume $\mathbf{C}$ by concatenating the corresponding unary features at different disparities between $[0, D-1]$. For the cost volume regularization, we apply a 3D convolutional neural network to regularize and transform the cost volume of size $D \times \frac{H}{4} \times \frac{W}{4} \times 2C$ into a probability volume $\mathbf{P}$ of size $D \times H \times W$. The disparity map is then regressed from the probability volume through the soft-argmin operation \cite{kendall2017end}. The $L_1$ loss between the disparity output and the ground truth disparity map will be calculated to train the network.

For detailed network architectures, we apply the standard 2D/3D UNet \cite{ronneberger2015u} structures to construct our 2D feature extraction network as well as the 3D cost volume regularization network. In later sections, we will introduce and validate the proposed stereo matchability framework based on this baseline network. But it is noteworthy that our method is naturally applicable to a large set of stereo networks such as PSMNet \cite{chang2018pyramid}. And in later section (Sec. \ref{sec:discussion}) we will show the effectiveness of our method on a lightweight regression network. 

\begin{figure*}[t!]
	\centering
	\includegraphics[width=1\linewidth]{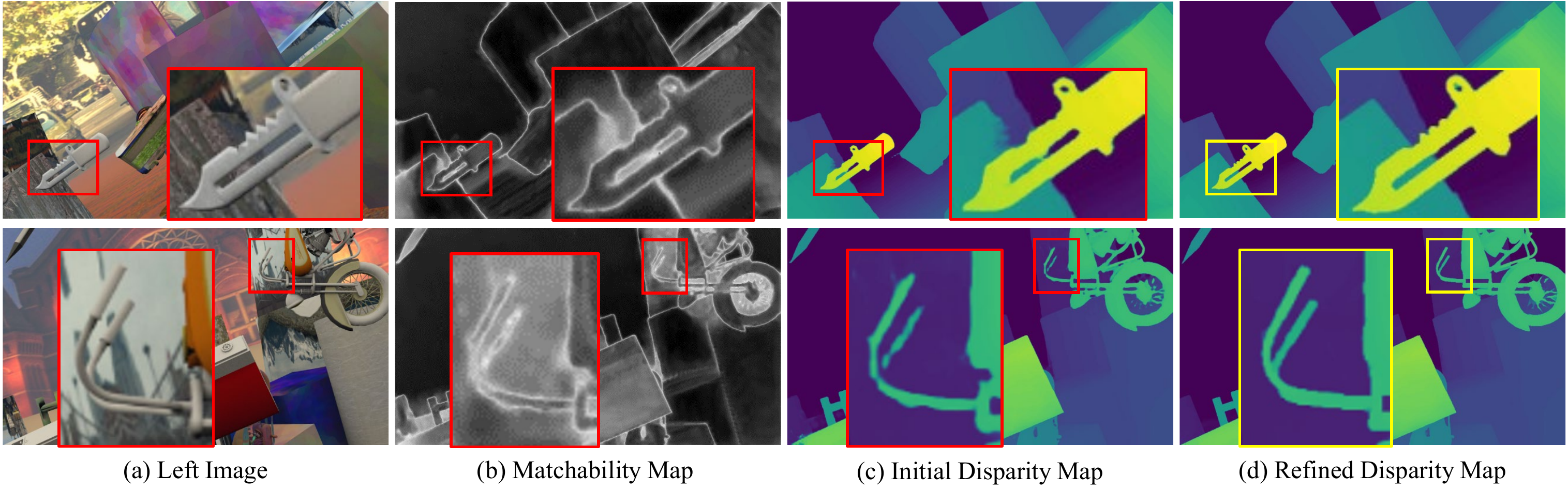}
	\vspace{-6mm}
	\caption{Illustrations on intermediate results of the proposed network. From left to right: (a) the left input image; (b) the regressed matchability map; (c) the initial disparity map; (d) the refined disparity map. These two samples clearly shows the effectiveness of the matchability-aware disparity refinement.}
	\label{fig:illustration}
	\vspace{-3mm}
\end{figure*}

\subsection{Matchability Regression from Probability Volume}\label{sec:entropy}

$\mathbf{P}(x, y, d)$ measures the probability that pixel $\mathbf{I}_l(x, y)$ in the left image is matched to pixel $\mathbf{I}_r(x - d, y)$ in the right image. For pixels that can be perfectly matched, the disparity-wise probability distribution should be uni-modal. We use the \textit{expectation} (also referred to \textit{soft-argmin} \cite{kendall2017end}) to approximate the best matching disparity:
\begin{equation}
\mathbf{D}(x, y) = \sum_{d=0}^{D-1} \mathbf{P}(x, y, d) \cdot d.
\end{equation}

On the other hand, the disparity-wise probability distribution also reflects the matching quality. Unmatchable pixels usually present multi-modal or random probability distributions. Based on this observation, we use the randomness of the distribution to measure the matchability. The \textit{entropy} operation is applied to directly regress the matchability from the probability volume:
\begin{equation}
\mathbf{M}(x, y) = \sum_{d=0}^{D-1} \mathbf{P}(x, y, d) \cdot \log\mathbf{P}(x, y, d).
\end{equation}
The entropy operation is differentiable and allows the joint training of the disparity and the matchability (Sec.~\ref{sec:loss}). Moreover, the matchability indicates where the disparity estimation is unreliable, which can be used as an important guidance to refine the disparity output. 

\subsection{Joint Disparity and Matchability Learning}\label{sec:loss}

Assuming that the observed disparity value $d_{gt}$ follows the Laplacian distribution \cite{kendall2017uncertainties}, we predict a mean disparity value $\hat{d}$ and a scale factor $\hat{b}$ (sometimes referred to output uncertainty) that maximize the likelihood: $p(d_{gt} | \hat{d}, \hat{b}) = 1 / (2\hat{b}) \cdot \exp{(|\hat{d} - d_{gt}| / \hat{b})}$. Let $\mathbf{B}$ denotes the scale map, $\mathbf{D}_{gt}$ denotes the ground truth disparity map. We minimize the negative log likelihood to jointly infer $\mathbf{D}$ and $\mathbf{B}$:
\begin{equation}\label{eq:likelihood}
\begin{split}
L_{joint} &= \sum_{x, y} -\log \big( \frac{1}{2\mathbf{B}(x, y)} \exp {\frac{|\mathbf{D}(x, y) - \mathbf{D}_{gt}(x, y)|}{\mathbf{B}(x, y)}} \big) \\
&= \sum_{x, y} \frac{|\mathbf{D}(x, y) - \mathbf{D}_{gt}(x, y)|}{\mathbf{B}(x, y)} + \log \mathbf{B}(x, y).
\end{split}
\end{equation}
Constants are ignored in the above equation. In practice, we instead infer $ \mathbf{B}'(x, y)=\log \mathbf{B}(x, y) $ so that $ \mathbf{B}(x, y)$ is implicitly enforced to be positive. 

For image-level vision tasks (e.g., monocular depth map estimation), it is straightforward to split the network at the last 2D convolutional layer to obtain both the output and the scale. However, in stereo reconstruction, $\mathbf{B}$ should be retrieved from the 3D probability volume. 
Notice that the matchability in Sec.~\ref{sec:entropy} also reflects the uncertainty of the observed disparity. We thus use a 2D CNN to transform the matchability map $\mathbf{M}$ to the uncertainty map $\mathbf{B}$ (Fig.\ref{fig:framework}). The loss formulation \ref{eq:likelihood} can also be interpreted as soft $L_1$ loss considering the pixel-wise matchability, where the significance of unmatchable pixels will be attenuated.

\subsection{Matchability-aware Disparity Refinement}\label{sec:refinement}

As is discussed in previous sections, the disparity map $\mathbf{D}$ can hardly be perfect for low-matchability regions. Also, the $L_{joint}$ loss relaxes unmatchable regions, which may lead to over-smoothed disparity boundaries. Usually, a refinement module is applied at the end of the network to fine-tune the output by exploiting input image semantics \cite{xu2017deep,yao2018mvsnet,cheng2019learning}. 

In our network, apart from refining $\mathbf{D}$ with the input image $\mathbf{I}_l$, we also use the matchability $\mathbf{M}$ as a guidance for the refinement. We apply the CSPN \cite{cheng2019learning} as our refinement module. Specifically, A 2D UNet structure is used to extract the per-pixel diffusion kernel $\boldsymbol{\kappa}_{k \times k}(x,y)$ from the concatenation of $\{\mathbf{D}, \mathbf{I}_l, \mathbf{M}\}$. Then, the initial disparity map $\textbf{D}$ is iteratively refined with the learned convolutional kernel map to generate the final disparity map output. As suggested by \cite{cheng2019learning}, we refine the disparity map for 24 iterations.

Compared with the refinement with only the input image and the initial disparity map, the matchability map provides direct information about where the disparity needs to be refined. In later experiments, we will show the advantage of introducing $\mathbf{M}$ to the disparity refinement. 

\subsection{Training Loss}\label{sec:total_loss}

The network architecture of the proposed framework is illustrated in Fig.~\ref{fig:framework}. During training, we consider three loss terms, including the $L_1$ loss of the initial disparity map, the disparity and matchability joint training loss $L_{joint}$ (Sec.~\ref{sec:loss}) and the $L_1$ loss of the refined disparity map.
\begin{equation}\label{eq:loss}
\begin{split}
L = \lambda_i \cdot L_1^{init} + \lambda_j \cdot L_{joint} + \lambda_r \cdot L_1^{ref}
\end{split}
\end{equation}
$L_1^{init}$ is the commonly used loss term for stereo regression, which is important for guarantee the quality of initial disparity map estimation; $L_{joint}$ is the joint training loss to balance the disparity and matchability estimations during training; $L_1^{ref}$ is final output loss to constrain the refined disparity map. The three terms are weighted using $\lambda_i$, $\lambda_j$ and $\lambda_r$ during training. We set $\lambda_i = \lambda_j = \lambda_r = 1$ in our experiments.

\begin{table*}[]
	\caption{Quantitative results on KITTI 2012 \& 2015 stereo benchmarks over non-occluded regions (Noc) and all pixels (All). The D1 error is the percentage of pixels with disparity error larger than 3 px and 5\% of the ground truth.}
	\resizebox{\textwidth}{!}{%
		\begin{tabular}{c|c c c|c c c|c c c|c c c}
			\specialrule{.2em}{.1em}{.1em}
			\multirow{3}{*}{Methods} & \multicolumn{6}{c|}{KITTI 2015} & \multicolumn{6}{c}{KITTI 2012} \\ \cline{2-13} 
			& \multicolumn{3}{c|}{Noc} & \multicolumn{3}{c|}{All} & \multicolumn{3}{c|}{Noc} & \multicolumn{3}{c}{All} \\ \cline{2-13} 
			& D1-bg & D1-fg & D1-all & D1-bg & D1-fg & \textbf{D1-all} & \textgreater{}2px & \textgreater{}3px & EPE & \textgreater{}2px & \textgreater{}3px & EPE \\ \hline
			DispNetC \cite{mayer2016large} & 4.11 \% & 3.72 \% & 4.05 \% & 4.32 \% & 4.41 \% & 4.34 \% & 7.38 \% & 4.11 \% & 0.9 px & 8.11 \% & 4.65 \% & 1.0 px \\ 
			MC-CNN \cite{zbontar2016stereo}  & 2.48 \% & 7.64 \% & 3.33 \% & 2.89 \% & 8.88 \% & 3.89 \% & 3.90 \% & 2.43\% & 0.7 px & 5.45 \% & 3.63 \% & 0.9 px \\
			GC-Net \cite{kendall2017end} & 2.02 \% & 5.58 \% & 2.61 \% & 2.21 \% & 6.16 \% & 2.87 \% & 2.71 \% & 1.77 \% & 0.6 px & 3.46 \% & 2.30 \% & 0.7 px \\ 
			PSMNet \cite{chang2018pyramid} & 1.71 \% & 4.31 \% & 2.14 \% & 1.86 \% & 4.62 \% & 2.32 \% & 2.44 \% & 1.49 \% & 0.5 px & 3.01 \% & 1.89 \% & 0.6 px \\ \hline
			\textbf{DSM} (Ours) & 1.66 \% & 4.16 \% & 2.07 \% & 1.83 \% & 4.56 \% & 2.28 \% & 2.25 \% & 1.39 \% & 0.5 px & 2.83 \% & 1.79 \% & 0.5 px \\ \hline
			SegStereo \cite{yang2018SegStereo} & 1.76 \% & 3.70 \% & 2.08 \% & 1.88 \% & 4.07 \% & 2.25 \% & 2.66 \% & 1.68 \% & 0.5 px & 3.19 \% & 2.03 \% & 0.6 px \\ 
			GwcNet \cite{guo2019group} & 1.61 \% & 3.49 \% & 1.92 \% & 1.74 \% & 3.93 \% & 2.11 \% & 2.16 \% & 1.32 \% & 0.5 px & 2.71 \% & 1.70 \% & 0.5 px \\
			EdgeStereo \cite{song2020edgestereo} & 1.69 \% & 2.94 \% & 1.89 \% & 1.84 \% & 3.30 \% & 2.08 \% & 2.32 \% & 1.46 \% & 0.4 px & 2.93 \% & 1.83 \% & 0.5 px \\
			GANet \cite{zhang2019ga} & 1.40 \% & 3.37 \% & 1.73 \% & 1.55 \% & 3.82 \% & 1.93 \% & 2.18 \% & 1.36 \% & 0.5 px & 2.79 \% & 1.80 \% & 0.5 px \\
			CSPN \cite{cheng2019learning} & 1.40 \% & 2.67 \% & 1.61 \% & 1.52 \% & 2.88 \% & 1.74 \% & 1.79 \% & 1.19 \% & -* & 2.27 \% & 1.53 \% & -* \\
			\specialrule{.2em}{.1em}{.1em}
		\end{tabular}
	}
	*Not reported by the paper or the benchmark
	\label{tab:kitti}
	\vspace{-3mm}
\end{table*}

\begin{table*}[]
	\caption{Quantitative results on Scene Flow dataset. The settings of different network architectures are described in Sec.~\ref{sec:ablation}. Corresponding network architectures are also visualized in Fig.~\ref{fig:ablation}}
	\resizebox{\textwidth}{!}{%
		\begin{tabular}{c|c|c|c|c|c c c}
			\specialrule{.2em}{.1em}{.1em}
			\multirow{2}{*}{Settings} & \multirow{2}{*}{Initial Loss} & \multirow{2}{*}{Matchability Loss} & \multicolumn{2}{c|}{Refinement Loss} & \multirow{2}{*}{EPE} & \multirow{2}{*}{\textgreater 1 px} & \multirow{2}{*}{\textgreater 3 px} \\ \cline{4-5}
			&  &  & w.  Matchability & w.  Input Image &  &  &  \\ \hline
			Baseline Network & \checkmark &  &  &  & 0.875 px & 9.07 \% & 4.30 \% \\ 
			Baseline w/ Refinement & \checkmark &  &  & \checkmark & 0.842 px & 9.20 \% & 4.24 \% \\ 
			Baseline w/ Matchability & \checkmark & \checkmark &  &  & 0.946 px & 9.25 \% & 5.06 \% \\ 
			Refinement w/o Matchability as Input & \checkmark & \checkmark &  & \checkmark & 0.794 px & 8.84 \% & 4.24 \% \\ 
			DSM w/o Initial Loss &  & \checkmark & \checkmark & \checkmark & 0.807 px & 8.65 \% & 4.28 \% \\ 
			\textbf{DSM} (proposed) & \checkmark & \checkmark & \checkmark & \checkmark & \textbf{0.761 px} & \textbf{8.31 \%} & \textbf{4.07 \%} \\ 
			\specialrule{.2em}{.1em}{.1em}
		\end{tabular}%
	}
	\vspace{-2mm}
	\label{tab:sceneflow}
\end{table*}

\begin{figure*}[]
	\centering
	\includegraphics[width=1\linewidth]{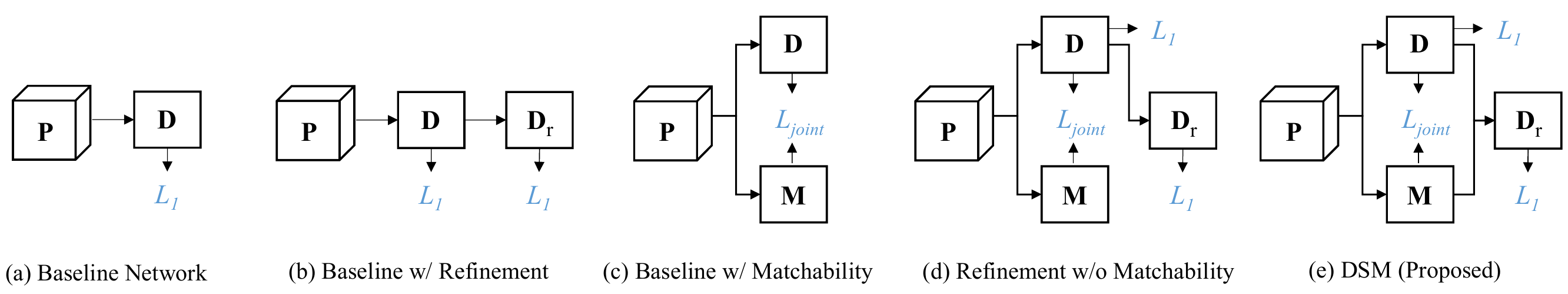}
	\vspace{-6mm}
	\caption{Visualizations on the five different network architectures discussed in ablation study section~\ref{sec:ablation}. Corresponding quantitative results are shown in Table~\ref{tab:sceneflow}}
	\label{fig:ablation}
	\vspace{-3mm}
\end{figure*}


\begin{figure*}
	\begin{subfigure}
		\centering
		\includegraphics[width=\linewidth]{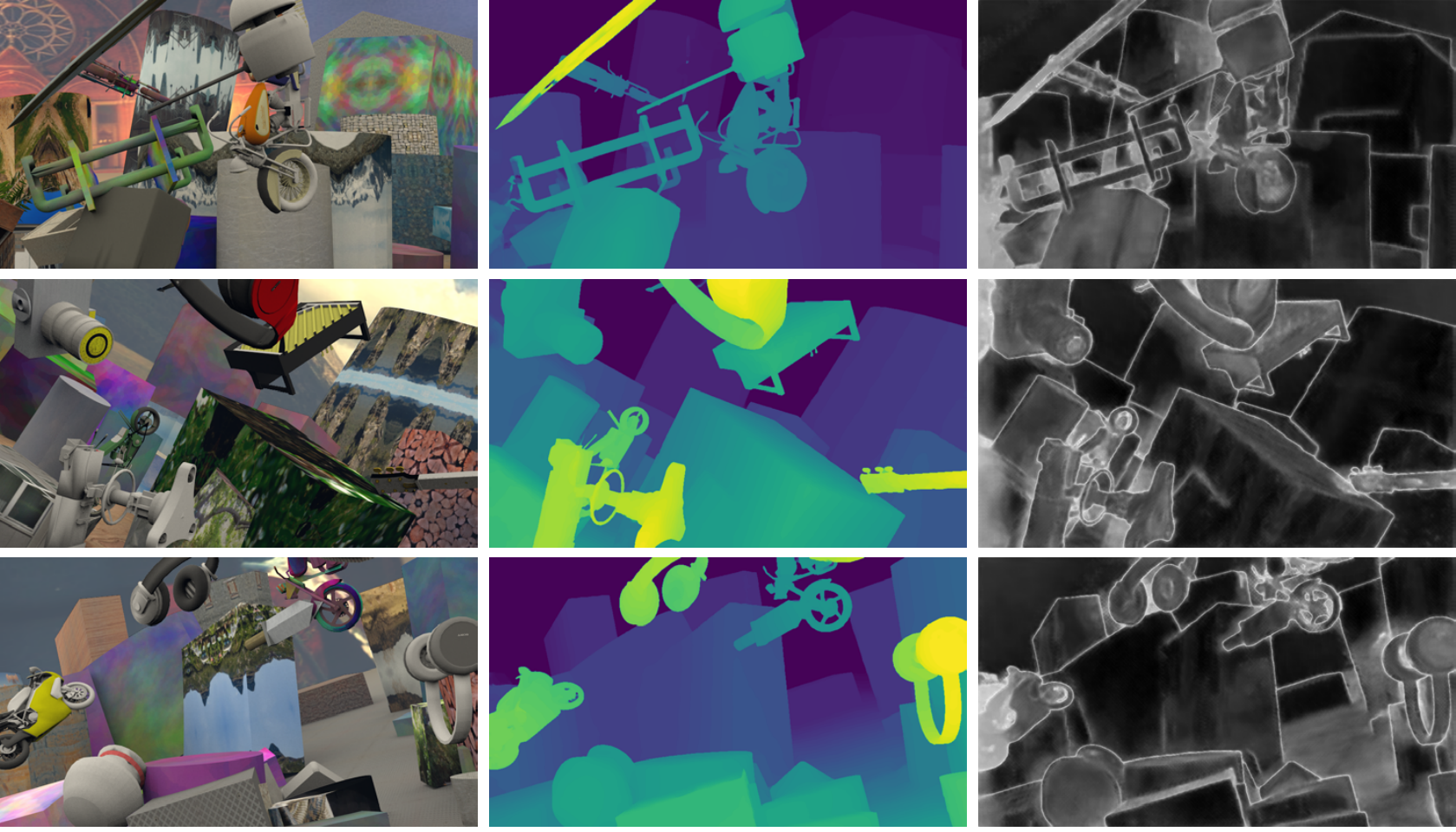}
		\vspace{-8mm}
		\caption{Results on Scene Flow dataset. From left to right are the left image, refined disparity map and the matchability map.}
		\label{fig:sceneflow}
	\end{subfigure}
	\vspace{4mm}
	\begin{subfigure}
		\centering
		\includegraphics[width=\linewidth]{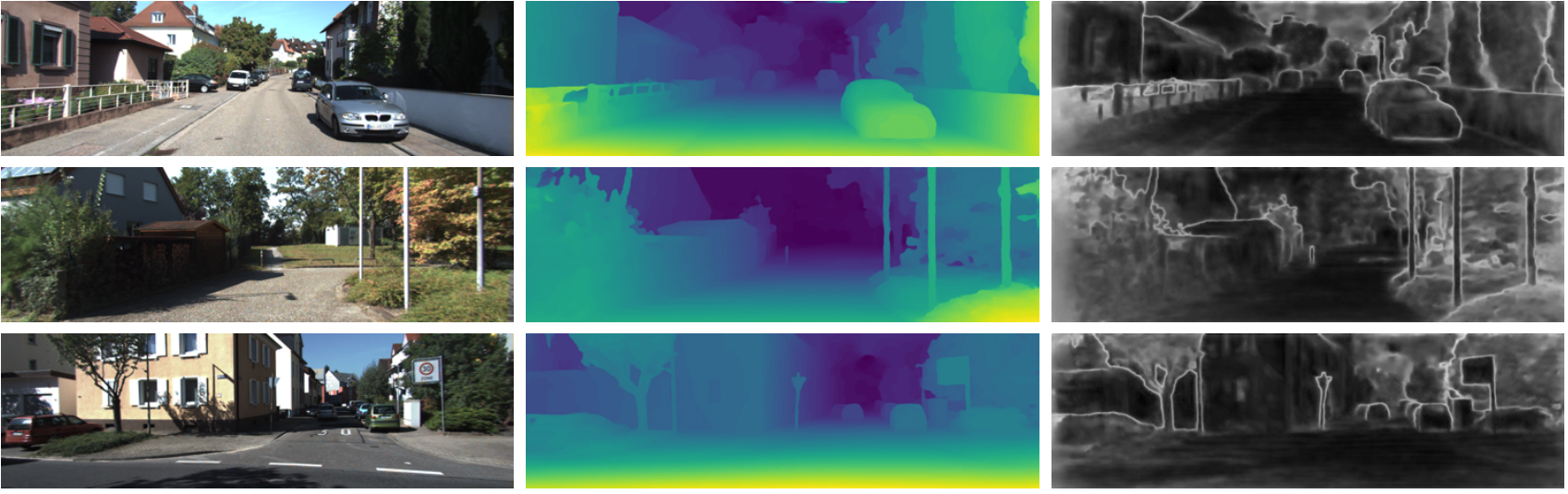}
		\vspace{-8mm}
		\caption{Results on KITTI 2012 benchmark. From left to right are the left image, refined disparity map and the matchability map.}
		\label{fig:kitti2012}
	\end{subfigure}
	\vspace{4mm}
	\begin{subfigure}
		\centering
		\includegraphics[width=\linewidth]{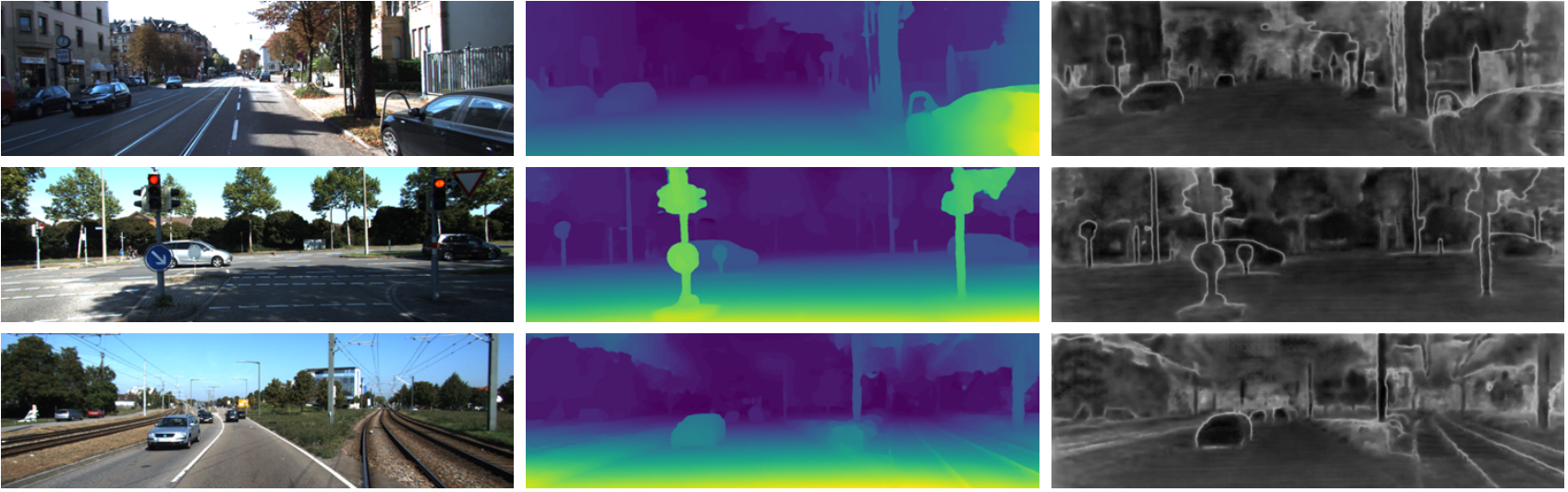}
		\vspace{-8mm}
		\caption{Results on KITTI 2015 benchmark. From left to right are the left image, refined disparity map and the matchability map.}
		\label{fig:kitti2015}
		\vspace{-5mm}
	\end{subfigure}
\end{figure*}

\section{Experiment}

\subsection{Datasets and Implementation Details}

We apply Adam optimizer with $\alpha=0.9$ and $\beta=0.999$ to train the model. All training samples are first intensity normalized and then randomly cropped to $H \times W = 256 \times 512$ before being fed into the network. When calculating the network loss, we exclude all pixels with disparity larger than 192, which is set as the max disparity in the practice. All our models are trained on four P100 GPUs with a batch size of 16 in total. The program take 15GB VRAM on each GPU. We first train our network on the Scene Flow dataset  \cite{mayer2016large} for totally 16 epochs. The initial learning rate is 0.001 and is halved after epoch 10, 12 and 14. Then, we fine-tune our model using the KITTI  \cite{geiger2012we,menze2015object} training data. The network is further trained for 600 epochs on KITTI dataset with a learning rate of 0.001. We decay the learning rate by 10 at epoch 200 and 400.

For testing, we set input image size to $H \times W = 544 \times 992$ for Scene Flow dataset and $H \times W = 384 \times 1248$ for KITTI dataset. The disparity output is then resized back to the original size for loss computation. When averaging the results over the Scene Flow test set, we exclude the samples whose proportion of valid pixels is less than $10\%$. 

\subsection{Evaluation Results}

\paragraph{KITTI Benchmarks}
We evaluate our methods on KITTI online benchmarks. For evaluation metrics, KITTI 2015 calculates the D1 error as the percentage of pixels with disparity error larger than 3 px and 5\% of the ground truth, while KITTI 2012 calculates the $>2$ px percentage and the $>3$ px percentage to benchmark all submissions. The quantitative results of our method are shown in Table \ref{tab:kitti}. DSM achieves error rates of $1.79\%$ on KITTI 2012 and $2.28\%$ on 2015 benchmark, outperforming previous stereo methods \cite{kendall2017end,chang2018pyramid}. Some methods \cite{song2020edgestereo,yang2018SegStereo} show better result on the KITTI 2015 but not on the KITTI 2012. The other listed methods \cite{guo2019group,zhang2019ga, cheng2019learning} surpass our method. 

For GANet and CSPN, one possible reason for their effectiveness is that well designed cost aggregation are applied. However, the aggregation methods are very diverse among the state-of-the-art stereo matching algorithms. Because this is not our focus, we choose plain 3D CNN as the regularization for generalization. It may raise concern that the proposed method utilizes CSPN module but produces worse result than \cite{cheng2019learning}. The reason is that we only use 2D CSPN for refinement, but do not use 3D CSPN in cost aggregation. 

\paragraph{Scene Flow Dataset} 
Three evaluation metrics are used to evaluate the results on Scene Flow dataset: 1) the end point error (EPE), which is the mean disparity difference among all pixels; 2) the percentage of pixels with disparity errors $>1$ pixel; 3) the percentage of pixels with disparity errors $>3$ pixels. As shown in Table~\ref{tab:sceneflow}, DSM achieves an EPE score of 0.761 px, which significantly reduces the end point error by 13 \% compared with our baseline network (0.875 px), showing the effectiveness of the proposed matchability learning framework.


\subsection{Ablation Studies}\label{sec:ablation}
In this section we study the effectiveness of each component in the proposed network. All models are trained and are tested in the Scene Flow dataset. Results are shown in Table~\ref{tab:sceneflow}.

\paragraph{Baseline Network} We fist evaluate our baseline network on Scene Flow dataset. Both the matchability regression and the matchability-aware disparity refinement are removed from the network (see Fig.~\ref{fig:ablation} (a)). As shown in Table \ref{tab:sceneflow}, the baseline network achieves an EPE of 0.87. In contrast, the EPE of the proposed framework is 0.77, which demonstrates the overall performance improvement by introducing the DSM framework.

\paragraph{Baseline with Refinement} Next, we add a disparity refinement module to the baseline network (see Fig.~\ref{fig:ablation} (b)). This setting is similar to previous methods \cite{xu2017deep,yao2018mvsnet,cheng2019learning} that we use the input image context as a guidance to fine-tune the disparity output. Both the initial disparity loss $L_1^{init}$ and the refined disparity loss $L_1^{ref}$ are considered during training. The EPE after the refinement is slightly reduced from 0.875 px to 0.842 px. 

\paragraph{Baseline with Matchability} In this setting we train the network with the joint training loss $L_{joint}$ described in Sec.~\ref{sec:loss} (see Fig.~\ref{fig:ablation} (c)). The network jointly learns the matchability and the disparity from training samples, and relaxes those pixels with low-matchability values during training. Adding the matchability alone results in slightly worse performance (0.946) than the baseline model. This is mainly because that the network is focused on matchable pixels during training, but during testing the errors of all the pixels are considered. If we evaluate matchable and unmatchable regions separately, we can find the disparity quality of matchable regions are improved. Here we intuitively consider a pixel as a matchable pixel if its attenuation weight is not down-scaled ($1 / \mathbf{B}(x, y) > 1$). Table \ref{discussion-match-table} shows that the loss attenuation improves the performance in matchable regions by a significant margin, while causing dramatic drop in unmatchable regions. 

\begin{table}
	\caption{Separated evaluations. We consider a pixel as matchable if its attenuation weight $1 / \mathbf{B}(x, y)$ is larger than 1.}	
	\resizebox{\linewidth}{!}{
		\begin{tabular}{c|c|c}	
			\specialrule{.18em}{.09em}{.09em}
			\multirow{2}{*}{Settings} & \multicolumn{2}{c}{EPE (px)} \\ \cline{2-3} 
			& Matchables & Unmatchables \\ \hline
			Baseline w/ Matchability & 0.140 & 6.623 \\
			Baseline Network & 0.212 & 5.515 \\ 		
			\specialrule{.18em}{.09em}{.09em}
		\end{tabular}
	}
	\label{discussion-match-table}
\end{table}

\begin{table}
	\caption{DSM with different implementations on baseline network and matchability regression method (Sec.~\ref{sec:discussion}). }
	\resizebox{\linewidth}{!}{%
		\begin{tabular}{l|c c c}		
			\specialrule{.18em}{.09em}{.09em}
			& EPE (px) & \textgreater 1 px (\%) & \textgreater 3 px (\%) \\ \hline
			Baseline & 0.875 & 9.07 & 4.30 \\ 
			DSM (proposed) & 0.761 & 8.31 & 4.07 \\ \hline
			\multicolumn{4}{c}{PSMNet* as Baseline} \\ \hline
			Baseline (PSMNet) & 0.857 & 9.11 & 4.05  \\ 
			DSM (PSMNet) & 0.740 & 8.18 & 3.85  \\ \hline
			\multicolumn{4}{c}{Matchability Regression Alternatives} \\ \hline
			DSM (2DCNN) & 0.769 & 8.43 & 4.09  \\ 
			DSM (3DCNN) & 0.867 & 10.03 & 4.29 \\
			\specialrule{.18em}{.09em}{.09em}
		\end{tabular}%
	}
	*The implementation of PSMNet is adopted from the released code.
	\vspace{-4mm}
	\label{tab:discussion}
\end{table}

\paragraph{DSM w/o Initial Loss} This setting is similar to the proposed DSM except that we do not compute the initial disparity loss $L_1^{init}$ during training. We observe from Table~\ref{tab:sceneflow} that the EPE score (0.807) is larger than the proposed DSM (0.761). 

This is mainly because $L_1^{init}$ is able to better constrain the initial disparity map, and the refinement module can produce improved output with better initial estimation. One of the common failures is that the disparity estimation in a certain region is completely wrong. These regions are also hard for refinement, given that the refinement module is strong at recovering the estimation according to image context and neighboring disparities instead of guessing without clue. By using $L_1^{init}$ to better constrain the initial disparity map, we can reduce such cases and improve the overall performance. 

\begin{table}
	\caption{Comparison of quality and running time between the lightweight model and other methods on Sceneflow test set. }
	\resizebox{\linewidth}{!}{
		\begin{tabular}{c|ccc|c}	
			\specialrule{.18em}{.09em}{.09em}
			Settings & EPE (px) & \textgreater1px (\%) & \textgreater3px (\%) & Time (s) \\ \hline
			Baseline & 0.875 & 9.07 & 4.30 & 0.32 \\ 
			DSM & 0.761 & 8.31 & 4.07 & 0.34 \\ 
			Baseline (lightweight) & 0.952 & 9.66 & 4.56 & 0.15 \\ 
			DSM (lightweight) & 0.806 & 8.75 & 4.08 & 0.17 \\
			\specialrule{.23em}{.09em}{.09em}
		\end{tabular}
	}	
	\vspace{-4mm}
	\label{tab:discussion-time}
\end{table}

\paragraph{Refinement w/o Matchability as Input} This setting is similar to the proposed DSM except that we do not pass the matchability map as input to the refinement network (see Fig.~\ref{fig:ablation} (d)). The network regresses the matchability map for robust initial disparity map estimation but does not utilize the matchability information for disparity refinement. As shown in Table \ref{tab:sceneflow}, the EPE is reduced from 0.875 px to 0.794 px. However, the error is still larger than with the matchability as input, which demonstrates the effectiveness of the matchability information for disparity refinement.

\subsection{Discussion}\label{sec:discussion}

\paragraph{Application to Existing Networks} The proposed matchability learning framework is applicable to most of the recent deep stereo networks \cite{kendall2017end,chang2018pyramid,cheng2019learning,zhang2019ga}. Except for the proposed UNet baseline network, here we also apply our method to the pyramid stereo matching network (PSMNet \cite{cheng2019learning}) and the result on Scene Flow dataset are shown in Table~\ref{tab:discussion}. PSMNet is slightly better than our UNet baseline. Also, after applying the DSM framework the EPE errors of both the UNet baseline and PSMNet are reduced by $\sim 0.1$, which shows that our method is able to consistently boost performances of such stereo networks.

\paragraph{Real-time Applications} We also replace the baseline model with a lightweight one and show that the proposed method can improve the quality by a significant margin while preserving the high efficiency. The number of 3D convolutional layers is decreased and the cost volume is constructed by the absolute difference of two feature vectors instead of concatenation. We compare the quality and inference time on Sceneflow dataset. The test images have the resolution of $H \times W = 540 \times 960$. As is shown in Table \ref{tab:discussion-time}, the EPE boosts from 0.952 (Baseline (lightweight)) to 0.806 (DSM (lightweight)) on the lightweight base model. Moreover, the lightweight model produces even better result than the full baseline with only half of the runtime, which demonstates its potential for real-time applications or platforms with limited computation power. It can achieve 20 fps with $H \times W = 320 \times 576$ input on GTX2080Ti. 

\paragraph{Matchability Regression Alternatives} 

Instead of explicitly retrieving the matchability, we attempt to learn the mapping from the probability volume to the uncertainty map. First we treat the disparity dimension of the probability volume as the feature channel and apply a 2D CNN to it.  Although this setting (DSM (2DCNN)) achieves better EPE, treating the disparity dimension as the feature channel requires a fixed disparity sample number $D$, which limits the generalization of the method. Second we apply a 3D CNN followed by average pooling over disparity dimension to probability volume. Due to memory limitation, the volume is downsized to a quarter, which also results in downsized uncertainty map. This setting (DSM (3DCNN)) degrades the EPE. 


\section{Conclusion}
\vspace{-0.5mm}
We have presented a joint disparity and matchability regression framework for learning-based stereo matching. We have proposed to use the disparity-wise \textit{entropy} operation to retrieve the matchability information from the probability volume, and have derived a robust loss function to reduce the negative influence of unmatchable pixels during training. Moreover, instead of refining the disparity map only with the input image, we further exploit the matchability map and take it as a guidance to refine the disparity output especially for weakly matchable regions. Our method has been extensively studied on Scene Flow dataset and KITTI benchmarks, demonstrating the effectiveness of each step of the proposed framework.



\section{Acknowledgments}
This work is supported by Hong Kong RGC GRF 16206819 \& 16203518 and T22-603/15N.




\vspace{-0.5mm}
\bibliographystyle{IEEEtran}
\bibliography{IEEEabrv,bare_conf}
%



\end{document}